\pdfoutput=1
\PassOptionsToPackage{prologue,dvipsnames}{xcolor}

\documentclass[11pt]{article}

\usepackage[preprint]{acl}

\usepackage{times}
\usepackage{latexsym}

\usepackage{kotex}
\usepackage{multirow}
\usepackage{makecell}

\usepackage[dvipsnames]{xcolor}
\PassOptionsToPackage{dvipsnames}{xcolor}
\usepackage{booktabs}
\usepackage{amsmath}
\usepackage{adjustbox}

\usepackage[T1]{fontenc}

\usepackage[utf8]{inputenc}

\usepackage{microtype}
\usepackage{graphicx}

\usepackage{inconsolata}

\title{K-Act2Emo: Korean Commonsense Knowledge Graph \\ 
for Indirect Emotional Expression}

\author{Kyuhee Kim, Surin Lee \and Sangah Lee \\
  Seoul National University \\
  \texttt{\{salgu, lsrwhite, sanalee\}@snu.ac.kr}}

\begin{document}
\maketitle
\begin{abstract}
In many literary texts, emotions are indirectly conveyed through descriptions of actions, facial expressions, and appearances, necessitating emotion inference for narrative understanding. In this paper, we introduce K-Act2Emo, a Korean commonsense knowledge graph (CSKG) comprising 1,900 indirect emotional expressions and the emotions inferable from them. We categorize reasoning types into inferences in positive situations, inferences in negative situations, and inferences when expressions do not serve as emotional cues. Unlike existing CSKGs, K-Act2Emo specializes in emotional contexts, and experimental results validate its effectiveness for training emotion inference models. Significantly, the BART-based knowledge model fine-tuned with K-Act2Emo outperforms various existing Korean large language models, achieving performance levels comparable to GPT-4 Turbo.
\end{abstract}

\section{Introduction}
Literary texts often convey characters' emotions indirectly through descriptions of gestures, facial expressions, or appearances. For example, we can infer someone's anxiety from their hand gestures. These nonverbal cues act as common knowledge within specific societies or cultures \citep{culturalsense}, allowing readers to interpret the emotional state of characters. However, datasets addressing this task have been confined to subsets of commonsense knowledge graphs (CSKGs) (\citealp{speerconceptnet};\citealp{atomic2019};  \citealp{zhang2020}, \citealp{atomic2020}). These often intertwine inferences on emotions with non-emotional elements, and their coverage is limited.

\begin{figure}[ht!]
\centering
\includegraphics[width=7.8cm]{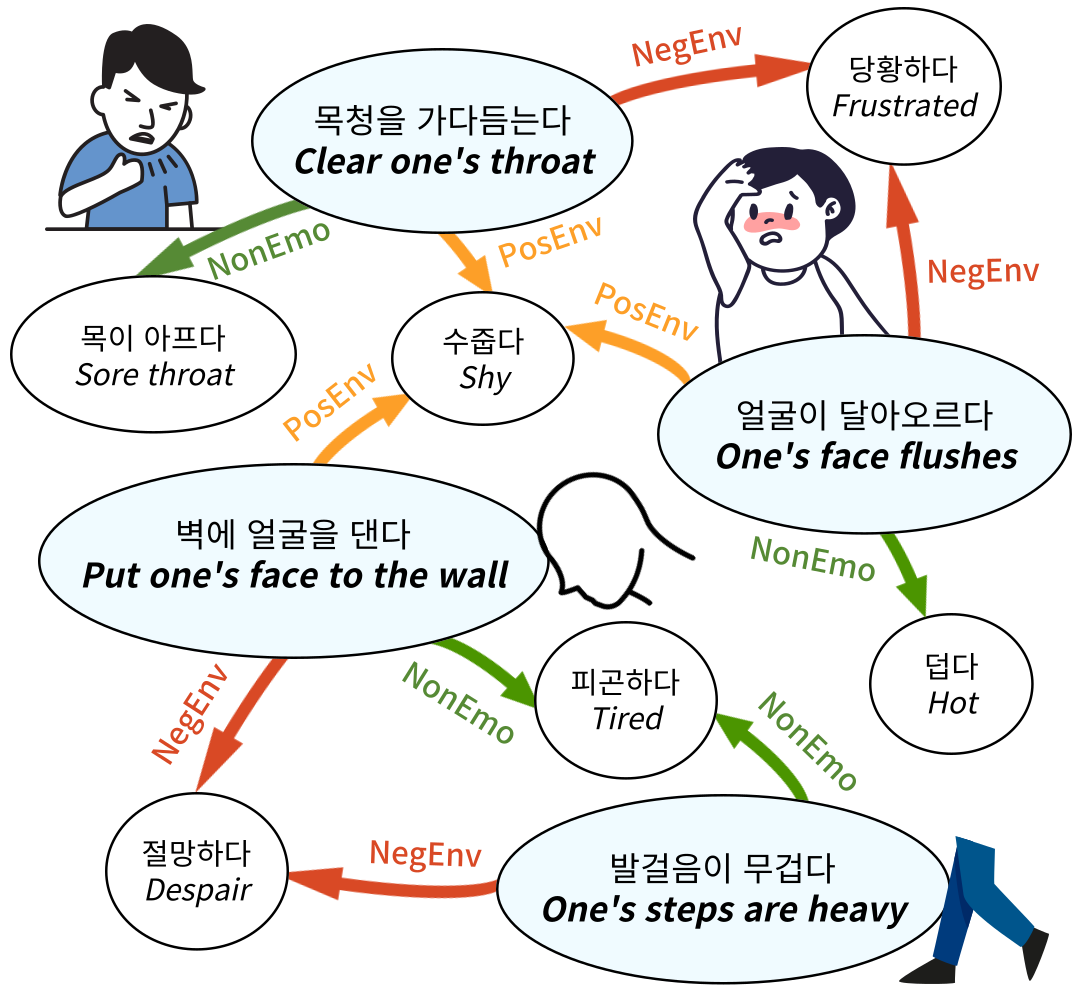}
\caption{Illustration of inferential knowledge in K-Act2Emo: \textit{PosEnv} for inferences in positive situations, \textit{NegEnv} for negative situations, and \textit{NonEmo} when expressions do not serve as emotional cues.}
\end{figure}

In this paper, we introduce K-Act2Emo\footnote{\label{sec:repo}We release our dataset and experiment results to the community at \url{https://github.com/koreankiwi99/K-Act2Emo.git}}, a Korean CSKG designed for indirect emotional expressions. K-Act2Emo comprises over 1,900 common indirect emotional expressions used by Koreans, with 6,002 inferred nodes included. We propose a new taxonomy for reasoning types: inference in positive situations (\textit{PosEnv}), in negative situations (\textit{NegEnv}), and when expressions do not serve as emotional cues (\textit{NonEmo}). This approach reflects that inference on the same action can vary depending on the context; for example, \textit{"His face turns red"} may imply either anger or amusement depending on the context. Using this taxonomy, we curate vast sets of triples for indirect emotional expression (Head), reasoning type (Relation), and inference (Tail) (see Table~\ref{tab:example}).

\begin{table*}[ht]
\centering
\resizebox{\textwidth}{!}{
\begin{tabular}{ccc}
\toprule
Head & Relation & Tail \\
\midrule
\multirow{6}{*}{
\makecell{바닥에서 구른다 \\ \textbf{\textit{Roll around on the floor}}}
}
& \colorbox{Yellow}{PosEnv}
& \makecell{유쾌하다, 재미있다 \\ \textbf{\textit{Joy, Fun}}}\\
\cmidrule{2-3}
& \colorbox{Melon}{NegEnv}
& \makecell{반항심을 가지다 \\ \textbf{\textit{Rebellious*}}}\\
\cmidrule{2-3}
& \colorbox{LimeGreen}{NonEmo}
& \makecell{아프다 \\ \textbf{\textit{Sick}}}\\
\bottomrule
\end{tabular}

\begin{tabular}{ccc}
\toprule
Head & Relation & Tail \\
\midrule
\multirow{3}{*}{
\makecell{목청을 가다듬는다 \\ \textbf{\textit{Clear one's throat}}}
}
& \colorbox{Yellow}{PosEnv}
& \makecell{기대하다, 긴장하다 \\ \textbf{\textit{Anticipating, Nervous*}}}\\
\cmidrule{2-3}
& \colorbox{Melon}{NegEnv}
& \makecell{불안하다, 당황하다 \\ \textbf{\textit{Anxious, Embarrassed}}}\\
\midrule
\makecell{이를 간다 \\ \textbf{\textit{Grind one's teeth}}}
& \colorbox{Melon}{NegEnv}
& \makecell{분노하다 \\ \textbf{\textit{Angry}}}\\
\bottomrule
\end{tabular}}
\caption{Example triple sets from K-Act2Emo. English translations are based on the closest meanings, though the usage contexts for words marked with an asterisk do not fully align.}
\label{tab:example}
\end{table*}

The collection of indirect emotional expressions from non-literary sources, such as news articles or online comments, presents significant challenges. The reliance on only publicly available literary works further constrains the diversity of expressions accessible. Additionally, the automatic extraction of inferences from corpora is prone to biases, potentially affecting the quality of the knowledge derived \citep{gordon2013reporting}. Therefore, in constructing K-Act2Emo, we employ a two-step process. First, we gather indirect emotional expressions through crowdsourcing, presenting the participants with a list of emotions and body parts for guidance. Then, we collect corresponding emotions and other inferences through a second round of crowdsourcing. This methodology enables us to assemble a corpus of indirect emotional expressions that are frequently used and considered natural by Koreans. It also allows us to obtain direct emotional expressions that intuitively align with the indirect expressions identified.

In our research with K-Act2Emo, we conduct the following studies: (1) We compare the coverage of an existing CSKG against that of K-Act2Emo. (2) We train a pre-trained language model (LM) on K-Act2Emo to investigate its impact on the model's performance. We establish that K-Act2Emo possesses more comprehensive knowledge for interpreting indirect emotional expressions, and the model fine-tuned with K-Act2Emo shows a marked improvement over the existing open-source Korean large language models (LLMs). According to both automatic and human evaluations, the performance of the LM trained with K-Act2Emo is comparable to that of GPT-4 Turbo, indicating significant advancements in the LM's ability to comprehend and interpret indirect emotional expressions.

\section{Background}
In the domain of natural language processing (NLP), indirect emotional expressions and the corresponding emotional inferences have been integrated into large-scale knowledge graphs (KGs). These KGs enhance the reasoning capabilities of language models, often in conjunction with neural representations (\citealp{speerconceptnet}; \citealp{atomic2019}; \citealp{comet}; \citealp{zhang2020}; \citealp{atomic2020}).

However, within the Korean context, there is a notable absence of comprehensive, high-quality Commonsense Knowledge Graphs (CSKGs), 
with the majority of available resources being translations from English counterparts. K-Act2Emo, developed with the active participation of Korean speakers, represents a relatively small yet indigenous CSKG. Given the absence of a comparable CSKG in Korean, we undertake a comparative analysis of K-Act2Emo's coverage against \textbf{\begin{math}\mathrm{ATOMIC}^{20}_{20}\end{math}}, a well-established, large-scale CSKG in English (refer to Section~\ref{sec:four}).

ATOMIC \citep{atomic2019} distinguishes itself from other Knowledge Graphs (KGs) such as CONCEPTNET (v5.7) \citep{speerconceptnet} and TRANSOMCS \citep{zhang2020} through its specialized focus on event-based inferences and social commonsense knowledge. This encompasses dynamic event features like causes and effects, if-then relationships, and mental states. It is structured with 877K triples in the \textbf{Head-Relation-Tail} format, across 9 reasoning types (Relations), each collected and validated via crowdsourcing. The subsequent extension, \text{\begin{math}
\mathrm{ATOMIC}^{20}_{20}
\end{math}}, broadens the scope with 1.33M triples across 23 reasoning types. K-Act2Emo adopts this triple format as well, but with a modification in the reasoning types to better accommodate inferences from indirect emotional expressions.

\begin{table}[h!]
\centering
\resizebox{7.8cm}{!}{
\begin{tabular}{lcc}
\toprule
\textbf{Relation} & \textbf{Description} & \textbf{Type}\\
\midrule
xReact & as a result, PersonX feels & Mental State\\
xIntent & because PersonX wanted & Mental State\\
xWant & as a result, PersonX wants & Event\\
xEffect &  as a result, PersonX will & Event\\
xAttribute & PersonX is seen as & Persona\\
\bottomrule
\end{tabular}}
\caption{Reasoning types in \begin{math}
\mathrm{ATOMIC}^{20}_{20}
\end{math} which are potentially related to emotions. \textit{PersonX} refers to the agent of the Head event.}
\label{tab:atomic}
\end{table}

Among the reasoning types in \begin{math}
\mathrm{ATOMIC}^{20}_{20}
\end{math}, those most relevant to K-Act2Emo include \textit{xReact} and \textit{xAttribute} (Table~\ref{tab:atomic}). However, inferences about emotions are distributed across others, such as \textit{xEffect} and \textit{xWant}. This dispersal complicates separating emotional from non-emotional inferences.

KGs, while beneficial, grapple with issues of scalability and expressiveness, lacking the flexibility inherent in Language Models (LMs). To address this gap, there is an increasing interest in research that combines KGs with LMs and explores the use of LMs as dynamic knowledge bases. COMET \citep{comet} demonstrates that pre-trained LMs can leverage commonsense knowledge graphs to generate new, valid inferences about events. \citet{west} propose Symbolic Knowledge Distillation, a machine-to-corpus-to-machine approach that employs large language models to augment commonsense knowledge graphs. This method trains smaller-sized language models on these enhanced commonsense knowledge graphs to produce high-quality causal commonsense knowledge. In this study, we also aim to assess transformer-based LMs' knowledge in emotional inference and the benefits of incorporating K-Act2Emo with LMs.

\section{K-Act2Emo}
\subsection{Collection}

\textbf{Indirect Emotional Expression Collection} 
Our initial step in collecting indirect emotional cues involved crowdsourcing from 10 Korean native speakers with extensive experience in creating literary works. We provided them with emotion-related words from \citet{hongjeong} and a comprehensive list of body parts (see Appendix~\ref{sec:appendix}). To avoid repetitive submissions, we made the growing corpus publicly available in real-time, rewarding only those submissions that were genuinely original compared to the existing corpus. We considered indirect emotional expressions as new if they exhibited significant differences in nuance. As generating novel expressions became increasingly difficult, we escalated the rewards and introduced guidelines to prevent direct emotional expressions, such as "wearing a sad smile." After the collection phase, we standardized verb tenses to present tense and unified personal references (e.g., friends, family, lovers) under the singular term "Y" to ensure consistency. \\

\begin{figure}[t!]
\centering
\includegraphics[width=8cm]{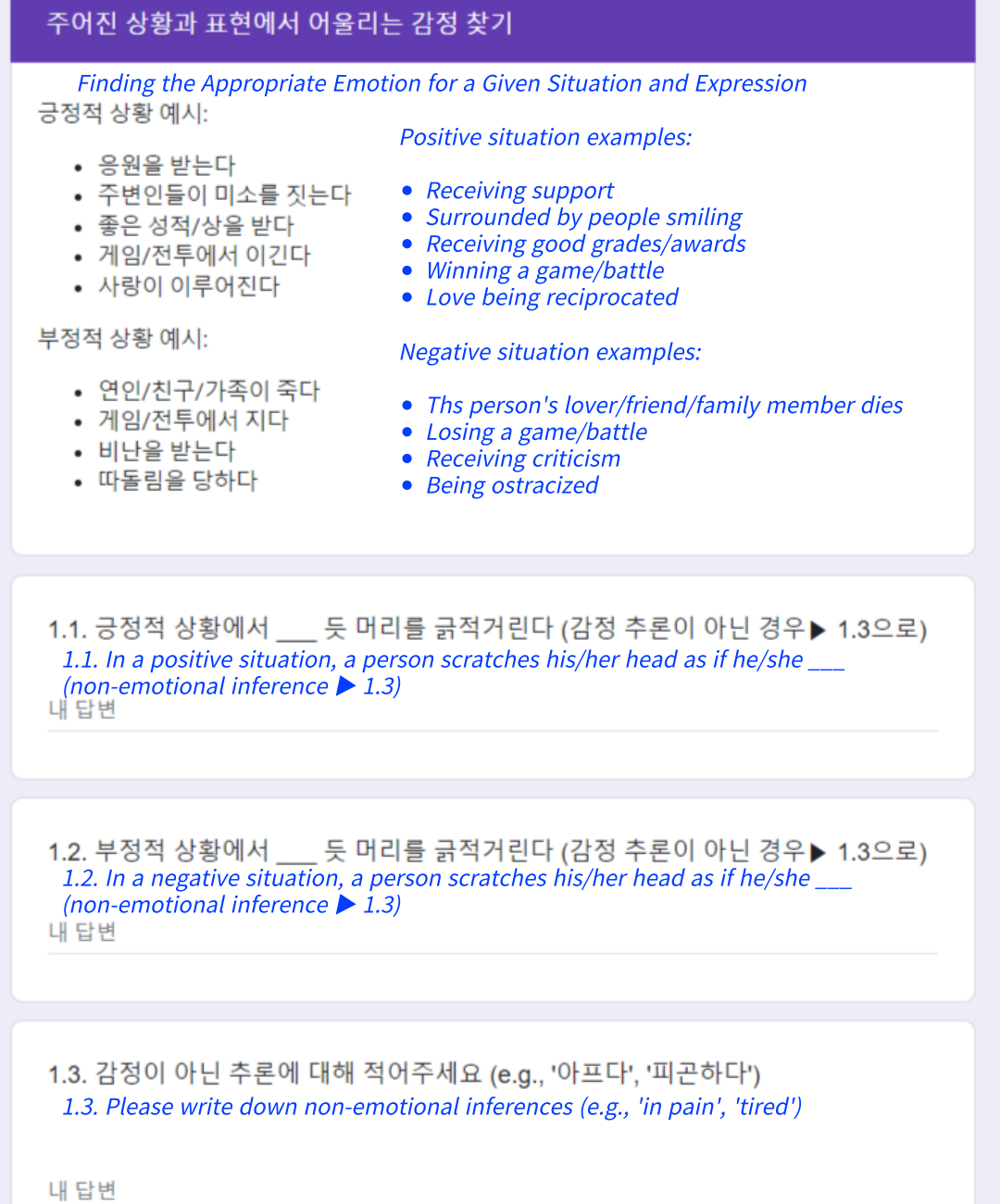}
\caption{Template of the crowdsourcing task for gathering inferences on indirect emotional expressions}
\label{fig:templ}
\end{figure}

\noindent\textbf{Reasoning Type Design}
Before gathering emotions from indirect expressions, we conducted two pilot tests to design proper reasoning types. The first phase, when we collected emotions without reasoning types, identified a wide range of emotions across various scenarios, often skewed by the possibility of multiple interpretations. To mitigate this, we classified situations into positive (\textit{PosEnv}) or negative (\textit{NegEnv}) categories, offering clear examples like "receiving support" or "a family member passing away" to direct participants towards more precise contexts. The second pilot, which involved collecting emotions with both positive and negative situational factors (\textit{PosEnv} and \textit{NegEnv}), resulted in consistent emotional inferences within an event; however, there were still issues with reporting non-emotional aspects, such as "being sick." To address these discrepancies, we introduced a new category, \textit{NonEmo} Relation, to more accurately document these inferences. \\

\noindent\textbf{Inference Collection} In order to gather emotional expressions inferred from indirect cues, we conducted another crowdsourcing session with 20 Korean speakers. To grasp nuanced emotions and expedite the process, we allowed participants to freely describe emotions using adjectives and phrases, rather than selecting from fixed emotional categories. We encouraged them to use direct emotional words over metaphors, limiting descriptions at most three words. Using the template "\_\_\_ deut (as if he/she feels \_\_\_)" along with the situation descriptions for Relation (see Figure~\ref{fig:templ}), participants filled in natural emotional expressions. Up to five inferences per indirect cue were permitted. If emotions inferred for both \textit{PosEnv} and \textit{NegEnv} contexts were identical, participants could select one. To avoid unnatural inferences, an option for \textit{No Inferred Emotion} was provided. Participants were free to enter any inferences unrelated to emotional expressions in the \textit{NonEmo} field.

\subsection{Statistics}
Similar to ATOMIC, the annotations in K-Act2Emo is a range of potential inferences. Therefore, to validate our dataset, we enlist 20 workers who were involved in the collection of inferences to indicate whether they agree with the contributions of other workers. In this process, we utilize a 4-point Likert scale (always/often, sometimes/likely, farfetched/never, invalid), used by \citet{atomic2020}, where the first two labels are considered \textbf{Accept} and the last two \textbf{Reject}. However, unlike \citet{atomic2020}, we do not provide a \textit{No judgement} option. The evaluation results are as follows: out of a total of 7,272 nodes, 6,002 were accepted, resulting in an acceptance rate of 82.53\%. The acceptance rate for nodes in \textit{NegEnv} is slightly lower compared to that in \textit{PosEnv} 
(Table~\ref{tab:consent}).

\begin{table}[ht!]
\centering
\resizebox{7.8cm}{!}{
\begin{tabular}{lcccc}
\toprule
& \textbf{Accept} & \textbf{Reject} & \textbf{Acceptance (\%)} & \textbf{Total} \\
\midrule
Nodes & 6,002 & 1,270 & \textbf{82.53} & 7272\\
\midrule
rel.PosEnv & 1957 & 281 & 87.17 & 2238\\
rel.NegEnv & 3157 & 977 & 76.36 & 4134\\
rel.NonEmo & 894 & 6 & 99.33 & 900\\
\bottomrule
\end{tabular}}
\caption{Percentage of node acceptance in raw annotations evaluated by crowdsourcing participants}
\label{tab:consent}
\end{table}

\noindent\textbf{Node Analysis by Relation}
Table~\ref{tab:node} details our knowledge graph, containing 6,002 nodes for 1,900 unique indirect emotions. It notes 634 nodes without positive context inferences and 135 without negative context inferences. Additionally, it details Tail distribution: 1,951 in PosEnv, 3,157 in NegEnv, and 894 in NonEmo, showing more inferences in NegEnv. There are 64, 94, and 124 unique Tails in PosEnv, NegEnv, and NonEmo, respectively, highlighting greater diversity in negative contexts. Except for "kancelhata" (\textit{Desperate}), "kincanghata" (\textit{Nervous}), "nollata" (\textit{Surprised}), "tanghwanghata" (\textit{Embarrassed}), and "pwukkulepta" (\textit{Shy/Ashamed}), there is no overlap between the Tails used in PosEnv and NegEnv. The terms "mwukilyekhata" (\textit{Helpless}), "casinkami issta" (\textit{Confident}), and "hokisimi sayngkita" (\textit{Curious}) are observed in both non-emotional and emotional Tails.\\

\begin{table}[h!]
\centering
\resizebox{7.8cm}{!}{
\begin{tabular}{lcc}
\toprule
 &\textbf{Count} & \textbf{Percentage (\%)}\\
\midrule
\textbf{Heads} & 1900 &  \\
\hspace{3mm} w/o PosEnv Nodes & 634 & 33.36\\
\hspace{3mm} w/o NegEnv Nodes & 135 & 7.1\\ 
\midrule
\textbf{Nodes} (unique Tails) & 6002 (274)\\
\hspace{3mm}rel.PosEnv & 1951 (64) & 32.5 \\
\hspace{3mm}rel.NegEnv & 3157 (94) & 52.59 \\
\hspace{3mm}rel.NonEmo & 894 (124) & 14.89 \\
\bottomrule
\end{tabular}}
\caption{Statistics of Heads and Tails in K-Act2Emo}
\label{tab:node}
\end{table}

\noindent\textbf{Word and Morpheme Counts in Heads and Tails}
Table~\ref{tab:morpheme} provides an analysis of word and morpheme distributions within Heads and Tails. Heads average 2.80 words (range: 1 to 7), and exhibit an average of 4.91 morphemes (range: 1 to 13) based on Mecab-ko morpheme analyzer\footnote{https://bitbucket.org/eunjeon/mecab-ko/src/master/} analysis, indicating brief structures. Tails are shorter, averaging 1.43 words (range: 1 to 3) and 3.17 morphemes (range: 1 to 7), suggesting even more concise constructions. Notably, Tails in negative contexts are shorter than those in positive ones.

\begin{table}[ht!]
\centering
\resizebox{7.8cm}{!}{
\begin{tabular}{lcccccc}
\toprule
 & \multicolumn{3}{c}{\textbf{Word Count}} & \multicolumn{3}{c}{\textbf{Morpheme Count}}\\
& Avg. & Max. & Min. & Avg. & Max. & Min. \\
\midrule
\textbf{Head} & 2.80 & 7 & 1 & 4.91 & 13 & 1\\
\midrule
\textbf{Tail} & 1.45 & 3 & 1 & 3.17 & 7 & 1\\
\hspace{3mm}rel.PosEnv & 1.34 & 3 &  1 & 3.20 & 6 & 2 \\
\hspace{3mm}rel.NegEnv & 1.05 &2 & 1 & 2.86 & 4 & 1 \\
\hspace{3mm}rel.NonEmo & 1.45 & 3 & 1 & 3.38 & 7 & 1 \\
\bottomrule
\end{tabular}}
\caption{Analysis of word and morpheme counts in Head and Tail.}
\label{tab:morpheme}
\end{table}

\begin{table*}[ht!]
\centering
\resizebox{\textwidth}{!}{
\begin{tabular}{lcc}
\toprule
\multicolumn{3}{c}{\textbf{PosEnv Nodes}} \\ \midrule
\textbf{Emotion Category} & \textbf{Count} &  \textbf{Percentage (\%)} \\
\midrule
Joy \small{기쁨} & 667 & 41.82 \\
Relaxation \small{이완} & 177 & 11.1 \\
Shame \small{수치심} & 165 & 10.34 \\
Surprise \small{놀람} & 119 & 7.46 \\
Fear \small{두려움} & 118 & 7.4 \\
Desire \small{욕구} & 112 & 7.02 \\
Impression \small{감동} & 94 & 5.89 \\
Acceptance \small{수용} & 78 & 4.89 \\
Complex Emotion \small{복합감정} & 42 & 2.63 \\
Gratitude \small{감사} & 17 & 1.07 \\
Sympathy \small{동정} & 5 &0.31 \\
Freshness \small{상쾌함} & 1 & 0.06 \\
\\
\\
\midrule
Total & 1595 & 100 \\
\bottomrule
\end{tabular}

\begin{tabular}{lcc}
\toprule
\multicolumn{3}{c}{\textbf{NegEnv Nodes}} \\ \midrule
\textbf{Emotion Category} & \textbf{Count} &  \textbf{Percentage (\%)} \\
\midrule
Fear \small{두려움} & 562 & 21.14 \\
Agony \small{고뇌} & 514 & 19.33 \\
Sadness \small{슬픔} & 373 & 14.03 \\
Disgust \small{혐오} & 349 & 13.13 \\
Anger \small{분노} & 321 & 12.07 \\
Surprise \small{놀람} & 198 & 7.45 \\
 Complex Emotion \small{복합감정} & 94 & 3.54 \\
Shame \small{수치심} & 56 & 2.11 \\
Emptiness \small{허무} & 52 & 1.96 \\
Desire \small{욕구} & 39 & 1.47 \\
Sympathy \small{동정} & 38 & 1.43 \\
Loneliness \small{고독} & 24 & 0.9 \\
Sorry/Apology \small{미안함} & 22 & 0.83 \\
Regret \small{후회} & 17 & 0.64 \\
\midrule
Total & 2659 & 100 \\
\bottomrule
\end{tabular}}
\caption{Emotion category distribution across inference nodes in PosEnv and NegEnv Relations}
\label{tab:cat}
\end{table*}

\noindent\textbf{Number of Emotions Inferred per Head}
While the collection of responses in a free-form manner allows for the capture of a wide range of emotional expressions, categorizing these expressions is essential for identifying similar emotional states. Therefore, we employ the Korean emotional adjective and verb dictionary by \citet{hongjeong} to classify the inferred emotions into 21 distinct categories.

Table~\ref{tab:cat} presents the distribution of emotion categories across emotional inference nodes. In the PosEnv Relation, the categorization process resulted in a consolidation to 1,595 nodes across 12 Emotion Categories. In contrast, in the NegEnv Relation, this process led to a reduction to 2,659 nodes, encompassing 14 Emotion Categories. Emotion Categories common to both Relations include \textit{Surprise, Sympathy, Fear, Complex Emotion, Shame, and Desire}.\footnote{\textit{Fear} and \textit{Shame} appear in \textit{PosEnv} nodes because "pwukkulepta" (\textit{Shy/Ashamed}) is classified as \textit{Shame}, and "kincanghata" (\textit{Nervous}) (with a neutral nuance in Korean) is classified as Fear.}

The number of Tails per Head is significant as it can indicate the complexity of interpreting the Head. Table~\ref{tab:freq} demonstrates this complexity through Head counts per the number of Tails and emotion categories. For Heads with three Tails, the counts were 148 in \textit{PosEnv} and 361 in \textit{NegEnv}. After categorizing Tails into emotion categories, these numbers reduced to 35 and 119 in \textit{PosEnv} and \textit{NegEnv}, respectively. Similarly, for Heads with two Tails, counts decreased from 389 to 259 in \textit{PosEnv} and from 670 to 656 in \textit{NegEnv}. This reduction may be due to oversimplification because of categorization, but it also suggests that several Heads have Tails with similar meanings. A significant observation is the predominance of Heads with three emotion categories in \textit{NegEnv} compared to \textit{PosEnv}. This phenomenon can be attributed to the higher arousal levels in negative contexts, which allow for simultaneous interpretations of Anger and Fear, contrasting with the distinct states of Joy and Relaxation in positive contexts.

\begin{table}[ht!]
\centering
\resizebox{7.8cm}{!}{
\begin{tabular}{ccc}
\toprule
& \multicolumn{2}{c}{\textbf{Head Count}} \\ Tails & PosEnv & NegEnv \\ 
\midrule
1 & 729 & 734 \\ 
2 & 389 & 670 \\ 
3 & 148 & 361 \\ 
0 & 634 & 135 \\ 
\bottomrule
\end{tabular}

\begin{tabular}{ccc}
\toprule
& \multicolumn{2}{c}{\textbf{Head Count}} \\ 
Categories & PosEnv & NegEnv \\ 
\midrule
1 & 972 & 990 \\ 
2 & 259 & 656 \\ 
3 & \textbf{35} & \textbf{119} \\ 
0 & 634 & 135 \\
\bottomrule
\end{tabular}}
\caption{Comparison of Head count based on number of Tails/Emotion categories in PosEnv and NegEnv.}
\label{tab:freq}
\end{table}

\section{Knowledge Coverage Comparison}
\label{sec:four}

In this study, we compare K-Act2Emo with \textbf{\begin{math}\mathrm{ATOMIC}^{20}_{20}\end{math}}. We translate K-Act2Emo’s indirect emotional expressions into English using DeepL\footnote{https://www.deepl.com/translator}, standardizing personal references to "PersonX" and replacing "Y" with "PersonY" for consistency.

Table~\ref{tab:compare} demonstrates that translating the Heads from K-Act2Emo results in 1,707 unique indirect emotional expressions, indicating a reduction from the original 1,900 due to nuances lost in the translation process. Among these, 161 expressions overlap with those in \textbf{\begin{math}\mathrm{ATOMIC}^{20}_{20}\end{math}}. Conversely, the translation of K-Act2Emo's Tails yields 101 direct emotional expressions, with 80 aligning with \textbf{\begin{math}\mathrm{ATOMIC}^{20}_{20}\end{math}}'s nodes related to \textit{xAttr}, \textit{xEffect}, \textit{xIntent}, \textit{xNeed}, \textit{xReact}, \textit{xWant}. In total, K-Act2Emo and \textbf{\begin{math}\mathrm{ATOMIC}^{20}_{20}\end{math}} share 109 Head-Tail pairs.

\begin{table}[ht!]
\centering
\resizebox{7.8cm}{!}{%
\begin{tabular}{lccc}
\toprule
 & \textbf{Count} & \textbf{Included} & \textbf{Percentage (\%)}\\
\midrule
Translated Heads & 1707 & 161 & 9.43\\
Translated Tails & 101 & 80 & 79.2 \\
\bottomrule
\end{tabular}}
\caption{Translated Heads and Tails of K-Act2Emo and its inclusion in 
\textbf{\begin{math}\mathrm{ATOMIC}^{20}_{20}\end{math}}}
\label{tab:compare}
\end{table}

Table~\ref{tab:exclusive} displays Heads unique to K-Act2Emo as well as those found in both K-Act2Emo and \textbf{\begin{math}\mathrm{ATOMIC}^{20}_{20}\end{math}}. Despite \textbf{\begin{math}\mathrm{ATOMIC}^{20}_{20}\end{math}}'s extensive coverage of daily events, it often fails to include common expressions that convey emotions. This omission stems not merely from cultural variances but from its lack of specialization in indirect emotional expression. This underscores K-Act2Emo's depth in understanding emotional expressions, distinguishing it from \textbf{\begin{math}\mathrm{ATOMIC}^{20}_{20}\end{math}}.

\begin{table}[ht!]
\centering
\resizebox{7.8cm}{!}{%
\begin{tabular}{cl}
\toprule
\multirow{3}{*}{\textbf{\makecell{Exclusive Heads\\ (K-Act2Emo)}}} & PersonX turns pale \\
& PersonX echoes PersonY's words \\
& PersonX admits PersonX's fault  \\
\midrule
\multirow{3}{*}{\textbf{\makecell{Shared Heads}}} & PersonX rolls PersonX's eyes \\
& PersonX drinks alcohol \\
& PersonX tilts PersonX's head to the side  \\
\bottomrule
\end{tabular}}
\caption{Comparison of Heads unique to K-Act2Emo and those found in both K-Act2Emo and Atomic2020}
\label{tab:exclusive}
\end{table}

\begin{table}[b!]
\centering
\resizebox{7.8cm}{!}{%
\begin{tabular}{ccl}
\toprule
\textbf{KG} & \textbf{Type} & \textbf{Instance} \\
\midrule
\multirow{4}{*}{K-Act2Emo} & Head & PersonX looks around slowly. \\\cmidrule{2-3}
&PosEnv&peaceful, relaxed, tense\\\cmidrule{2-3}
&NegEnv&aware, tense, nervous\\\cmidrule{2-3}
&NonEmo&cautious\\
\midrule
\multirow{3}{*}{\textbf{\begin{math}\mathrm{ATOMIC}^{20}_{20}\end{math}}} & Head & PersonX glances around the room. \\ \cmidrule{2-3}
& xReact & aware, attentive, tired, relieved, relaxed\\ \cmidrule{2-3}
& xAttribute & aware, concerned, cautious, curious, etc.\\
\bottomrule
\end{tabular}}
\caption{Comparison of Heads unique to K-Act2Emo and those found in both K-Act2Emo and Atomic2020}
\label{tab:better}
\end{table}

Even when addressing similar Heads and Tails, K-Act2Emo's reasoning taxonomy proves to be more effective than that of \textbf{\begin{math}\mathrm{ATOMIC}^{20}_{20}\end{math}}. As demonstrated in Table~\ref{tab:better}, K-Act2Emo distinctly categorizes content into \textit{PosEnv}, \textit{NegEnv}, and \textit{NonEmo}. In contrast, \textbf{\begin{math}\mathrm{ATOMIC}^{20}_{20}\end{math}} not only redundantly repeats information across \textit{xAttr} and \textit{xReact} but also merges contents of disparate characteristics, complicating the analysis and application of inferences. This distinction highlights K-Act2Emo's refined taxonomy in managing emotional expressions, offering a more structured framework for interpreting complex emotional cues.

\begin{table*}[ht!]
\centering
\begin{tabular}{llcccccc}
\toprule
& & \textbf{Bleu-1} & \textbf{Bleu-2} & \textbf{Bleu-3} & \textbf{Bleu-4} & \textbf{ROUGE-L} & \textbf{BERT Score} \\ 
\midrule
\multicolumn{2}{c}{COMET-BART}  & 0.558 & 0.432 & 0.294 & 0.259 & 0.566 & 0.770 \\
\midrule
\multirow{6}{*}{1-shot} & GPT-4 Turbo &
\textbf{0.408} & \textbf{0.275} & \textbf{0.156} & 0.000 & \textbf{0.429} & 0.695 \\
& KoAlpaca & 0.105 & 0.037 & 0.010 & 0.000 & 0.186 & 0.647 \\
& KoPlatypus & 0.171 & 0.087 & 0.028 & 0.000 & 0.266 & 0.685\\
& Kullm & 0.156 & 0.084 & 0.031 & 0.000 & 0.211 & 0.686 \\
& Mi:dm & 0.365 & 0.242 & 0.115 & 0.000 & 0.372 & \textbf{0.732} \\
&ChatSKKU & 0.138 & 0.071 & 0.022 & 0.000 & 0.196 & 0.696\\

\midrule
\multirow{5}{*}{5-shot} & KoAlpaca  & 
0.276 & 0.162 & 0.056 & 0.000 & 0.365 & 0.708 \\
& KoPlatypus & 0.345 & 0.209 & 0.071 & 0.000 & 0.404 & 0.726
\\
& Kullm & 0.307 & 0.184 & 0.058 & 0.000 & 0.331 & 0.716\\
& Mi:dm & \textbf{0.409} & \textbf{0.276} & 0.112 & 0.000 & \textbf{0.436} & \textbf{0.744} \\ 	
& ChatSKKU & 0.370 & 0.241 & \textbf{0.114} & \textbf{0.049} & 0.395 & 0.721\\
\bottomrule
\end{tabular}
\caption{Results of the automatic evaluation for emotional inference on the K-Act2Emo test set.}
\label{tab:auto_eval}
\end{table*}

\section{Emotional Inference Experiment}
\noindent\textbf{Experiment Setup} In this study, we assess whether K-Act2Emo can be utilized for training emotional inference knowledge in Transformer-based models. We implement a hold-out evaluation strategy on the K-Act2Emo dataset, dividing nodes related to PosEnv and NegEnv into training, validation, and testing partitions comprising 4,124, 482, and 502 nodes respectively. These subsets include 1,520, 180, and 180 unique heads, ensuring no overlap of heads between them. We also examine the distribution of emotion categories across these subsets to avoid bias (see Appendix ~\ref{sec:d} for more details).

We fine-tune COMET-BART, a BART-based \citep{lewis-etal-2020-bart} model using the COMET framework \citep{comet}, with the K-Act2Emo training set to generate Tails given its Head and Relation. We use KoBART\footnote{https://github.com/SKT-AI/KoBART} as the base model. As a baseline, we compare it to a one-shot GPT-4 Turbo (0125-preview), using a single randomly sampled fact from the training dataset (with the same Relation as the testing fact). To further examine the emotional inference capabilities of prominent Korean LLMs, we conduct one-shot and five-shot learning with Mi:dm \citep{kt}, ChatSKKU\footnote{https://huggingface.co/jojo0217/ChatSKKU5.8B}, Kullm \citep{kullm}, KoAlpaca
\footnote{https://huggingface.co/beomi/KoAlpaca-Polyglot-5.8B}, and KoPlatypus\footnote{https://huggingface.co/kyujinpy/KoT-platypus2-7B}, using facts sampled from the training dataset for each. We choose one-shot over zero-shot learning because the performance in zero-shot learning is significantly lower, and it generates overly lengthy results compared to K-Act2Emo's Tails. For details on each training session, refer to Appendix ~\ref{sec:e}. \\

\noindent\textbf{Evaluation Setup} The evaluation of these models includes both automatic and human assessments. For automatic evaluations, we utilize BLEU \citep{bleu}, ROUGE-L \citep{rouge}, and BERT Score \citep{bertscore} metrics, with the tails for the test set as gold labels. We use the Mecab-ko morpheme analyzer as a tokenizer for evaluation, and for the BERT Score, we utilize KoBERTScore\footnote{https://github.com/lovit/KoBERTScore.git} based on the Kcbert \citep{kcbert} model. Human evaluations were conducted by five reviewers on the outputs from COMET-BART, one-shot GPT-4 turbo, and five-shot Mi:dm. They evaluated the acceptability of these models based on the criteria established by \citet{atomic2020}. \\

\begin{table}[t!]
\centering
\resizebox{7.8cm}{!}{
\begin{tabular}{llcccccc}
\toprule
& & \textbf{Malformed} & \textbf{Lengthy} & \textbf{No Response} \\
\midrule
\multirow{6}{*}{1-shot} & GPT-4 Turbo & 11 & 6 & 0 \\
& KoAlpaca & 101 & 221 & 22\\
& KoPlatypus & 88 & 31 & 2\\
& Kullm & 151 & 76 & 11\\
& Mi:dm & 57 & 9 & 0\\
&ChatSKKU & 268 & 92 & 72 \\

\midrule
\multirow{5}{*}{5-shot} & KoAlpaca & 57 & 51 & 5\\
& KoPlatypus & 17 & 2 & 4\\
& Kullm & 73 & 13 & 1 \\
& Mi:dm & 24 & 2 & 0\\ 	
& ChatSKKU & 30	& 6 & 1\\
\bottomrule
\end{tabular}}
\caption{Count of Malformed Responses, Excessively Long Responses, and No Responses. Responses not ending in "\textit{-da}" (sentence-ending particle in Korean) are classified as Malformed, and those over three words as Excessively Long. Fewer instances indicate better alignment with the demonstration.}
\label{tab:format_eval}
\end{table}

\noindent\textbf{Results} We present our principal findings in Table~\ref{tab:auto_eval}, 
Table~\ref{tab:format_eval}, and Table~\ref{tab:human_eval_final}. Table~\ref{tab:auto_eval} demonstrates that COMET-BART, trained with K-Act2Emo, outperforms other models across all automatic metrics.\footnote{Due to the overall brevity of the target Tails, the results for BLEU-4 were low.} A significant improvement is observed in all models when employing five-shot learning over one-shot learning. Besides COMET-BART, GPT-4 Turbo and Mi:dm emerge as the top performers. As detailed in Table~\ref{tab:format_eval}, these models demonstrate a high degree of conformity to the response format of the provided demonstrations. Remarkably, despite Mi:dm's low performance in zero-shot learning trials in the pilot test, it shows rapid adaptation after a single example. Table~\ref{tab:format_eval} illustrates an improvement in the response format with five-shot learning compared to one-shot learning across most of the models.

\begin{table}[t!]
\centering
\resizebox{7.8cm}{!}{%
\begin{tabular}{lccc}
\toprule
\textbf{Model} & \textbf{Accept} & \textbf{Reject} & \textbf{No Judgement} \\
\midrule
COMET-BART & 97.21 & 2.79 & 0\\
GPT-4 turbo (1-shot) & 95.41 & 4.59 & 0 \\
Mi:dm (5-shot) & 92.03 & 7.97 & 0 \\ 
\bottomrule
\end{tabular}}
\caption{Results of the human evaluation for emotional inference on the K-Act2Emo test set.}
\label{tab:human_eval_final}
\end{table}

In human evaluations, consistent performance differences are observed (see Table~\ref{tab:human_eval_final}). GPT-4 turbo one-shot and Mi:dm five-shot performances are rated 1.8 percentage points and 5.18 percentage points lower, respectively, compared to COMET-BART. These disparities are attributed to the inclusion of content related to characters' personalities and states rather than emotions (see Table~\ref{tab:midm_example}). Additionally, several grammatical errors are observed in the outcomes of Mi:dm. Although NonEmo-related nodes are not utilized in the experiments, these results suggest a necessity for leveraging such nodes to enhance alignment and reranking of generated responses.

\begin{table}[h!]
\centering
\resizebox{7.5cm}{!}{%
\begin{tabular}{ccc}
\toprule
\multicolumn{3}{c}{\textbf{Mi:dm}} \\
\midrule
\textbf{Head} &  \textbf{Relation} &  \textbf{Tail} \\
\midrule
몸에 힘을 준다 & \multirow{2}{*}{PosEnv} & 강하다 \\
Tense muscles & & Strong\\
\midrule
대화에 참여한다 & \multirow{2}{*}{PosEnv} & 귀 기울여 듣다  \\
Participate in the conversation & & Listen attentively \\
\midrule
옷을 잡아 뜯는다 & \multirow{2}{*}{NegEnv} & 굶주리다 \\
Tear clothes & & Starving \\
\midrule
입이 살짝 벌어져 있다 & \multirow{2}{*}{PosEnv} & 즐겁고 기대 있다* \\
One's mouth is slightly open & & Happy and expectant* \\
\midrule
\multicolumn{3}{c}{\textbf{GPT-4 Turbo}} \\
\midrule
\textbf{Head} &  \textbf{Relation} &  \textbf{Tail} \\
\midrule
입을 문지른다 & \multirow{2}{*}{PosEnv} & 추워한다 \\
Rub his/her lips & & Feel cold\\
\midrule
손을 들어올린다 & \multirow{2}{*}{NegEnv} & 도움을 요청하다 \\
Raise his/her hand & & Ask for help \\
\bottomrule
\end{tabular}}
\caption{Rejected Nodes from Outputs of Mi:dm and GPT-4. The sentence marked with an asterisk is grammatically awkward.}
\label{tab:midm_example}
\end{table}

\section{Discussion}
While our novel taxonomy of reasoning types demonstrate efficacy in distinguishing between positive and negative valences through \textit{PosEnv} and \textit{NegEnv}, it  has limitations, such as difficulty differentiating between emotions with the same arousal level, such as fear and anger. In light of these findings, we propose several critical areas for in-depth investigation to enhance the applicability of our knowledge graph.\\

\noindent\textbf{The Observer’s Perspective} All descriptive statements about others are inherently influenced by the observer’s perspective. It is crucial, however, to distinguish between statements heavily influenced by subjective viewpoints and those that are not. For example, the statement "PersonX is pretending to be sick" carries an implicit bias, suggesting that Person X is not genuinely ill—a presumption that cannot entirely negate the possibility of Person X’s sickness. Making this differentiation is crucial for accurate inference from descriptions. \\

\noindent\textbf{The Level of Volition in Actions} The link between emotions and actions, especially in if-then reasoning, is complex. For instance, although it might make sense to say, "PersonX laughs loudly because he feels angry," deducing that the laughter is due to anger defies typical logic. This highlights how voluntary actions can skew the true representation of emotions. Our study sought to keep emotional inferences within common-sense bounds, opting for "as if he/she feels" to prevent misassociating actions with emotions, like laughter with anger. This highlights the necessity for finer distinctions in subsequent analyses. \\

\noindent\textbf{Indicators of Stimuli} Consider the statement "PersonX hides Y behind him/her," where the implied emotion of fear attributed to PersonX does not necessarily extend to "Y". Given that emotions arise in response to external stimuli, the emotional reaction to different stimuli can vary within the same context. Our study did not specifically address sentences involving direct stimuli due to their infrequency. However, the analysis of indirect emotional expressions, akin to those employed in Aspect-Based Sentiment Analysis, necessitates a distinct analysis for each stimulus involved.

\section{Conclusion}
In conclusion, our research presents K-Act2Emo, a detailed Korean Commonsense Knowledge Graph designed for analyzing indirect emotional expressions in literature. With 1900 descriptions of actions, facial expressions, and appearances, plus 6002 inferences, K-Act2Emo notably improves the precision of emotional inference beyond current models. Our evaluations confirm K-Act2Emo's utility in boosting the efficacy of language models for emotional inference tasks. This introduction of K-Act2Emo provides a novel tool for delving into emotional contexts. It not only enhances language models' ability to parse complex emotional narratives but also paves the way for further innovations in understanding human emotions through text.

\bibliography{act2emo}

\appendix

\section{Materials for Collection of Indirect Emotional Expressions}

The list of emotion-related words from \citet{hongjeong} is available on our repository~\ref{sec:repo}.

\label{sec:appendix}
\begin{table}[h!]
\centering
\resizebox{7.8cm}{!}{
\begin{tabular}{lll}
\toprule
\textbf{Category} & \textbf{SubCategory} & \textbf{Keywords} \\
\midrule
전신 (Whole body) &  & 몸, 피부, 자세 (body, skin, posture) \\
\midrule
\multirow{7}{*}{얼굴 (Face)} & \multirow{2}{*}{눈 (Eyes)} & 눈, 눈가, 눈썹, 미간, 눈꼬리 \\
& & (eyes, eye area, eyebrows, glabella, tail of the eye) \\
\cmidrule{2-3}
& 코 (Nose) & 코, 콧구멍 (nose, nostrils) \\
\cmidrule{2-3}
& \multirow{2}{*}{입 (Mouth)} & 입, 입술, 치아, 이, 이빨, 입꼬리 \\
& & (mouth, lips, teeth, corners of the mouth) \\
\cmidrule{2-3}
& 이마 (Forehead) & 이마, 관자놀이 (forehead, temples) \\
\cmidrule{2-3}
& 귀 (Ears) & 귀, 귓볼, 귓바퀴 (ears, earlobe, helix) \\ \cmidrule{2-3}
& \multirow{2}{*}{하관 (Lower face)} & 하관, 턱, 볼, 뺨, 광대, 보조개\\
& & (lower face, chin, cheeks, cheekbones, dimples) \\
\midrule
\multirow{2}{*}{발화 (Speech)} &  & 목소리, 어조, 말투 \\
&&(voice, tone, manner of speaking) \\
\midrule
\multirow{9}{*}{상체 (Upper body)} & 머리 (Head) & 머리, 고개 (head, nod) \\ 
\cmidrule{2-3}
& \multirow{2}{*}{목 (Neck)} & 목, 목젖, 목울대, 뒷목 \\
&&(neck, Adam's apple, nape) \\
\cmidrule{2-3}
& \multirow{2}{*}{어깨 (Shoulder)} & 어깨, 쇄골, 날개뼈\\ & & (shoulder, collarbone, shoulder blade) \\
\cmidrule{2-3}
& 허리 (Waist) & 허리, 척추 (waist, spine) \\
\cmidrule{2-3}
& \multirow{2}{*}{팔 (Arm)} & 팔, 팔뚝, 팔꿈치, 겨드랑이\\ & & (arm, forearm, elbow, armpit) \\
\cmidrule{2-3}
& \multirow{2}{*}{손 (Hand)} & 주먹, 손, 손목, 손바닥, 손가락, 손톱 \\
& & (fist, hand, wrist, palm, fingers, fingernails) \\
\cmidrule{2-3}
& \multirow{2}{*}{가슴 (Chest)} & 가슴, 갈비뼈, 흉통, 심장, 폐 \\
&&(chest, ribs, breastbone, heart, lungs) \\
\cmidrule{2-3}
& \multirow{2}{*}{배 (Stomach)} & 위장, 배, 속, 명치 \\
&&(stomach, belly, gut, solar plexus) \\
\midrule
\multirow{4}{*}{하체 (Lower body)} & \multirow{2}{*}{다리 (Leg)} & 다리, 허벅지, 종아리, 정강이, 무릎 \\
&&(leg, thigh, calf, shin, knee) \\ \cmidrule{2-3}
 & \multirow{2}{*}{발 (Foot)} & 발, 발바닥, 발톱, 발가락, 뒷꿈치, 발목 \\
&& (foot, sole, toenail, toes, heel, ankle)\\
\bottomrule
\end{tabular}}
\caption{List of body parts used for collecting indirect emotional expressions.}
\end{table}

\section{Training Details}
\label{sec:e}

\noindent\textbf{Details about COMET-BART Training} For COMET-BART training, we utilize the Hugging Face version (hyunwoongko/kobart) of KoBART as a base model. We fine-tune BART for six epochs using a batch size of 32 and a learning rate of 3e-4 on an NVIDIA A100 GPU 40 GB. \\

\noindent\textbf{Details about LLM Generation} For both five-shot and one-shot learning, we utilize the following list of Korean LLMs and GPT-4 Turbo (0125-preview). We set the temperature and top-p sampling rate to 1.0 each. For the GPT-4 Turbo model, we follow the default hyperparameters suggested by the OpenAI Platform, except for the temperature, which we set to 0.7. \\

\noindent\textbf{List of Korean Large Language Models} \\

\noindent\textbf{Mi:dm} is a 5.7B-sized, instruction-tuned auto-regressive language model. It is trained on both Korean and English in-house dataset. We utilize the Hugging Face version (KT-AI/midm-bitext-S-7B-inst-v1).\\

\noindent\textbf{ChatSKKU} is a 5.8B-sized, instruction-tuned model based on polyglot-ko-5.8b \citep{polyglotko} (a Korean Large Language Model trained with 172 billion tokens using the GPT-NeoX framework). It is instruction-tuned with the SKKU dataset, which includes a daily Korean conversation dataset. We utilize the Hugging Face version (jojo0217/ChatSKKU5.8B). \\

\noindent\textbf{KoAlpaca} We use a 5.8B-sized version of KoAlpaca based on polyglot-ko-5.8. The KoAlpaca Dataset v1.1b, consisting of a translated version of the Stanford Alpaca dataset and ChatGPT-generated data based on the Naver Q\&A dataset, is used for instruction-tuning. We utilize the Hugging Face version (beomi/KoAlpaca-Polyglot-5.8B). \\

\noindent\textbf{KoPlatypus} We use the 7B-sized COT version of KoPlatypus, based on the LLaMA 2 \citep{llama2} transformer architecture. The KoCoT\_2000 dataset, a translated version of the KAIST-CoT dataset, is used for instruction-tuning. We utilize the Hugging Face version (kyujinpy/KoT-platypus2-7B). \\

\noindent\textbf{Kullm} We use a 5.8B-sized version of Kullm based on polyglot-ko-5.8b. The Kullm v2 dataset, which includes translated versions of the GPT-4 LLM, Vicuna, and Dolly datasets, is used for instruction-tuning. We utilize the Hugging Face version (nlpai-lab/kullm-polyglot-5.8b-v2). \\

\noindent\textbf{Templates used in few-shot learning} \\

\noindent Table~\ref{tab:template} is a basic version of the templates we use for one-shot and five-shot learning. "Positive" can change to "Negative" in the case of NegEnv, and the example is randomly sampled from the training set each time. We adapt the template to the usage guideline of each model. For example, when we use Mi:dm, we replace “Response:” with “Midm;”. 

\begin{table}[ht!]
\centering
\resizebox{7.8cm}{!}{%
\begin{tabular}{l}
\toprule
\textbf{Template}\\
\midrule
아래는 작업을 설명하는 명령어와 추가적 맥락을 제공하는 입력이 \\짝을 이루는 예제입니다. \\
요청을 적절히 완료하는 응답을 작성하세요.\\
\\
\#\#\#명령어: 긍정적인 상황 속에서 이 사람의 감정을 추론하라.\\
한 문장으로 답하라.\\
\\
\#\#\#입력 1: Y에게 가까이 다가간다. \\
\#\#\#응답 1: 친밀감을 느끼다.\\
\\
\#\#\#입력 2: 고개를 갸웃거린다.\\
\#\#\#응답 2: \\
\midrule
Below is an instruction that describes a task, paired with an input that \\provides further context. \\
Write a response that appropriately completes the request.\\
\\
\#\#\#Instruction: 
Infer this person's emotion in a positive situation.\\ Answer in one sentence.\\
\\
\#\#\#Input 1: He/she approaches Y closely. \\
\#\#\#Response 1: He/she feels intimacy. \\
\\
\#\#\#Input 2: He/she tilts his/her head.\\
\#\#\#Response 2: \\
\bottomrule
\end{tabular}}
\caption{Templates used in one-shot learning. For five shot learning, we add four more examples.}
\label{tab:template}
\end{table}

\section{Statistics on Emotion Categories of Data Split}

\label{sec:d}
\begin{table}[ht]
\centering
\resizebox{7.8cm}{!}{
\begin{tabular}{lccc}
\toprule
\textbf{Emotion Category} & \textbf{Train (\%)} & \textbf{Valid (\%)} & \textbf{Test (\%)} \\
\midrule
두려움 (Fear) & 15.88 & 15.85 & 16.98 \\
기쁨 (Joy) & 15.88 & 13.66 & 16.05 \\
고뇌 (Agony) & 11.83 & 13.41 & 12.79 \\
슬픔 (Sadness) & 8.64 & 9.76 & 8.84 \\
혐오 (Disgust) & 8.20 & 6.59 & 9.77 \\
분노 (Anger) & 7.91 & 7.56 & 4.65 \\
놀람 (Surprise) & 7.38 & 8.54 & 6.98 \\
욕구 (Desire) & 3.13 & 6.10 & 4.42 \\
이완 (Relaxation) & 4.16 & 4.39 & 3.95 \\
수치심 (Shame) & 5.51 & 3.90 & 3.95 \\
복합감정 (Complex Emotion) & 3.40 & 2.68 & 2.09 \\
허무 (Emptiness) & 1.03 & 1.71 & 2.33 \\
감동 (Impression) & 2.28 & 1.46 & 2.33 \\
수용 (Acceptance) & 1.87 & 1.46 & 1.86 \\
동정 (Sympathy) & 0.97 & 0.98 & 1.40 \\
후회 (Regret) & 0.32 & 0.24 & 1.16 \\
미안함 (Sorry/Apology) & 0.50 & 0.98 & 0.23 \\
감사 (Gratitude) & 0.47 & 0.24 & - \\
상쾌함 (Freshness) & - & - & 0.23 \\
\bottomrule
\end{tabular}}
\caption{Emotion Categories of Train, Valid, and Test Splits}
\label{table:emotion_categories_comparison_compact}
\end{table}
\end{document}